# A New 3D Segmentation Methodology for Lumbar Vertebral Bodies for the Measurement of BMD and Geometry

André Mastmeyer*, Klaus Engelke*, WA Kalender

Institute of Medical Physics, Erlangen, Germany

**Abstract.** In this paper a new technique is presented that extracts the geometry of lumbar vertebral bodies from spiral CT scans. Our new multi-step segmentation approach yields highly accurate and precise measurement of the bone mineral density (BMD) in different volumes of interest which are defined relative to a local anatomical coordinate systems. The approach also enables the analysis of the geometry of the relevant vertebrae. Intra- and inter operator precision for segmentation, BMD measurement and position of the coordinate system are below 1.5% in patient data, accuracy errors are below 1.5% for BMD and below 4% for volume in phantom data. The long-term goal of the approach is to improve fracture prediction in osteoporosis.

## 1 Introduction

**Aim.** Low BMD a very important risk factor for osteoporotic fractures of the femur and the spine. Quantitative computed tomography (QCT) is one of the techniques to analyze trabecular BMD of the vertebrae L1-L3. Tradionally single slices of the mid-vertebral sections were scanned and analyzed [1]. Based on an anatomy oriented coordinate system trabecular and cortical VOI were defined. In this contribution we present a new three-dimensional approach using spiral CT scans that enables a volumetric analysis of BMD and an assessment of the vertebral geometry.

**State of the Art.** Lang et al. [2] achieved precision errors of 1.3 % for volumetric BMD measurements at the spine. However, they did not feature a full 3D segmentation of the vertebral bodies, which is a prerequisite in this context. A semi-automatic approach facing this issue has been recently presented by Kaminsky et. al in [3]. Before labelling they first manually separated the individual vertebral levels [4] which took 1-2 hours. More methodology-focused techniques used statistical shape models. Leventon [5] employed a combination of statistical models and Level Sets. Statistical models in the segmentation field still show deficits in locations with high curvature, which may be attributed to insufficient training data. The problem of the generation of suitable statistical shape models of the spine is also addressed in Vrtovec [6].

The segmentation of bony structures for QCT has been extensively explored in our group recently [4]. A combination of local thresholds adapted to the local noise environment and morphological hole filling works quite successfully, although some user interaction is still required. In the contribution presented here we extended our previous work to the spine and replaced the initial segmentation step by a deformable model. In the literature these are categorized as explicit models using meshes [7] or implicit formulations also called Level Sets [8]. To avoid computational costs and delicate numerical problems associated with the Level Sets we have chosen an explicit three-dimensional triangle mesh.

## 2 Methods and Materials

The proposed scheme works from coarse to fine segmentation. First domain-specific pre-segmentation constraints, which are described in more detail below, are determined. Then a deformable balloon surface is iteratively moved towards the outer cortical shell on radially emerging profiles. In the last step the segmentation in locations with high curvature is improved by local volume growing.

**Constraints.** Initially the user has to mark the center of the vertebral bodies. In a second step the center of the spinal channel is detected by a rolling ball procedure similar to Kaminsky [3]. Then planes are automatically fitted into the inter-vertebral disks. Based on the landmarks defined in these three steps each vertebral body is enclosed in an irregular box built of a boolean combination of a cylinder and the planes that limit the search space in the following segmentation step. The boxes effectively exclude the aorta, the vertebral processes and the adjacent vertebral bodies.

---

*corresponding authors {andre.mastmeyer, klaus.engelke}@imp.uni-erlangen.de

**Balloon Segmentation.** The balloon model chosen here consists of a content-adaptive triangle mesh with automatic regularization. From a theoretical point of view the algorithm solves the Euler-Lagrangian equations of motion. They are coupled to a minimization problem of internal and external energy of the balloon. The internal energy depends on the connectivity forces, the external energy is derived from the grey values of the image. The periosteal edges of the cortical shell attract the vertices of the balloon while the internal forces try to pertain a smooth and energy minimizing shape of a sphere. As a consequence the procedure is able to overcome difficulties due to noise and can bridge gaps caused by noisy spurious contours or low contrast. Practically the problem is solved by integration of the equations of motions

$$m\ddot{\vec{p}} + \gamma\dot{\vec{p}} = \vec{f}_{bal} = \vec{f}_{inf} + \vec{f}_{smg} + \vec{f}_{img}$$

for every vertex at position $\vec{p}$ using finite differences [7]. $\vec{f}_{bal}$ is the sum of the inflation forces $\vec{f}_{inf}$, external image forces $\vec{f}_{img}$, and smoothing forces $\vec{f}_{smg}$ that determine the viscosity of the balloon surface. In our implementation the damping $\gamma$ and the inflation forces are ignored. An exclusion of the latter prevents the well known "leaking out" problem. If no attractive points are present close to the surface of the balloon its surface points only move due to the influence of the viscosity term. In order to improve the local flexibility of the balloon surface new vertices are added in stretched areas. Here additional degrees of freedom are needed so that the surface can adjust itself to fine structures of the anatomy. In the vertebrae this is the case once the balloon surface flows into the edges close to the endplates. The external image forces are determined along profiles that radially emerge from the balloon surface [7]. For each vertex the grey values along the normals are sampled with sub-voxel precision from the image. Due to the smoothness term that is a major regularizing part of a deformable model the surface is not able to catch all points in highly curved regions. The smoothing forces are modelled by spring forces between vertices and their neighbours.

**Multiseeded Volume Growing.** To include the missing surface points mentioned above surface points with high BMD are searched on the balloon surface that pass the following threshold criteria: A low and a high threshold are set around the intersection of two Gaussian distributions using a positive and negative empirical offset. They are fitted to the typically bimodal histogram of the grey values in the search space of one vertebra. Voxels below and above both thresholds are labeled as soft tissue and bone respectively. In the transition zone between the two thresholds, a local noise adaptive criterion is used to separate the two classes. Voxels satisfying the threshold criteria for bone on the surface are used as seeds for local volume growing operations. These are followed by a closing and hole-filling procedure described in [4]. During this step parts of the processes are also included in the segmentation. These have to be cut-off now in a reproducible and robust manner without any user influence.

**Pedicle Cut.** In order to precisely determine the volume of the vertebral body the processes must be removed by an automatic procedure. In a nutshell, we try to find the smallest dissection surface through the pedicles between the vertebral body and the transverse processes (see Figure 1).

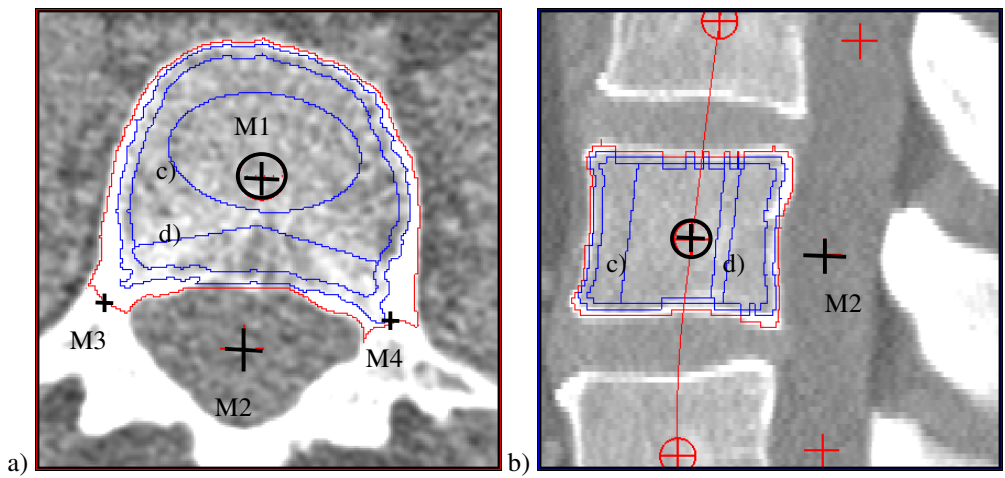

**Figure 1:** a) axial view of L2 from a patient at 60 mAs with attached markers, b) sagittal view of L2, neighboring levels and spline medial axis, c) cylindrical VOI, d) 'pacman' VOI

The methods used here are derived from the morphological concepts of ultimate erosion and Skeleton by Influence Zones (SKIZ), which are related to the ideas of Euclidean Distance Transforms and Voronoi partition of space [9]. First the segmentation result is eroded to its main components, the residuals of the vertebral body and the process. Then non-intersecting parallel dilations of the residuals are carried out. The two contact areas of the dilated residuals define the desired dissection.

**Trabecular Compartment.** A grey value based locally adaptive erosion is used to remove voxels with high grey values from the cortical shell of the vertebral body. The high threshold described above is used for this purpose. In a second step a further homogeneous erosion is used to peel off the sub cortical bone in order to assess the pure trabecular bone of the vertebral body which should improve precision of the trabecular BMD measurements because the sub cortical zone is excluded from the trabecular VOI.

**Vertebral Coordinate System and VOI generation.** Four landmarks are used for the definition of a vertebral coordinate system (VCS). The first is the centre of volume (COV) of the vertebral body (see Figure 1a, M1). The central line of the vertebral column is approximated by an interpolating spline curve of the COVs of all segmented vertebral bodies. The second landmark (M2) is defined as the intersection of a plane that is perpendicular to the spline curve with the center line of the spinal channel. Landmarks three and four (M3, M4) are defined as centers of the dissection areas between vertebral body and transverse processes (see above). The origin of the VCS is identical to the first landmark, the COV of the vertebral body. The x-axis is defined by the vector pointing toward M2, the z-axis is the tangent to the spline curve (Figure 1b) and the y-axis is perpendicular to the other two. The VCS is used to reproducibly position a cylinder and a so called 'pacman' VOI which also uses M3 and M4 for the placement of the exluded segment in the area of the basivertebral veins (Figure 1c,d).

**Performance Evaluation: Accuracy** was assessed using the European Spine Phantom, a geometrically defined, semi-anthropomorphic phantom [10]. CT datasets were obtained on a Siemens Sensation 16 (Siemens Medical Solutions, Forchheim, Germany) at 120 kV and a slice thickness of 1 mm. In order to investigate the influence of noise the radiation dose was varied: three different time current products (580, 145 and 36 mAs) were used. Relative to 580 mAs the other two settings increase noise by factors of 2 and 4. Each dataset was analyzed three times by the same operator. BMD and volume were determined in the trabecular compartments. Results were averaged and compared to the nominal BMD values of the phantom. For volume measurement the subcortical bone was included, for the BMD measurement it was excluded. BMD was analyzed for the total trabecular segmentation, the cylindrical and the 'pacman' VOIs.

**Intra- and inter-operator precision** was analyzed using clinical routine abdominal scans from 10 patients. CT acquisition was again performed on a Siemens Sensation 16 (60 mAs, 120 kV, slice thickness 1 mm). For each patient datasets with three different fields of view (FOV) (150, 250, 350 mm) with corresponding in plane pixel sizes of 0.3, 0.5 and 0.7 mm were used. For intra-operator analysis all datasets were analyzed three times by the same operator, for inter-operator analysis all datasets were analyzed once by three operators. Precision errors were determined for BMD and volume of the vertebral bodies L1-L3 as follows. For each patient and each of the three analyses results for L1-L3 were averaged. Then all 10 patients were averaged resulting in a percent coefficient of variation (%CV) per analysis. Finally root mean square averages were calculated for the three analyses resulting in an average %CV and a corresponding standard deviation SD. Afterwards inter-operator precision was determined using the same calculations only for the FOV with best results from the previous analysis (150 mm).

## 3 Results

| FOV [mm] | | 150 | | | 250 | | | 350 | | |
|---|---|---|---|---|---|---|---|---|---|---|
| Time current product [mAs] | | 580 | 145 | 36 | 580 | 145 | 36 | 580 | 145 | 36 |
| L2 | BMD1 | 1,10 | 0,91 | 0,44 | 0,52 | 0,51 | 0,23 | 1,08 | 0,84 | 0,25 |
| | BMD2 | 1,38 | 1,95 | 0,90 | 1,16 | 1,75 | 0,16 | 1,08 | 1,46 | 0,75 |
| | BMD3 | 1,27 | 1,02 | 0,06 | 0,67 | 0,64 | 0,70 | 1,22 | 0,97 | 0,35 |
| L1 | Vol. | 0,38 | 0,42 | 0,96 | 0,05 | 0,70 | 2,20 | 0,25 | 1,31 | 3,43 |
| L2 | Vol. | 3,48 | 0,31 | 1,20 | 3,14 | 0,09 | 0,17 | 1,96 | 0,03 | 0,17 |
| L3 | Vol. | 2,18 | 0,12 | 2,73 | 2,17 | 2,57 | 4,51 | 2,27 | 2,71 | 3,52 |

**Table 1: Accuracy errors in % from nominal value. BMD1: total trabecular VOI, BMD2: cylindrical VOI, BMD3: 'Pacman' VOI, Vol.: total trabecular volume.**

**Accuracy.** As seen in Figure 1 the proposed multi-step procedure accurately determined outer cortical and inner segmentation. The trabecular and geometrically defined VOIs analyzed in this study are shown as inner areas embedded in the cut-off outer total segmentation (Figure 1a). As can be seen most accuracy errors are below 1.3% for BMD and below 4% for volume. The higher errors for volume can be explained by an inclusion of a small part of the transverse processes into the segmentation of the vertebral body. This is unproblematic in patients.

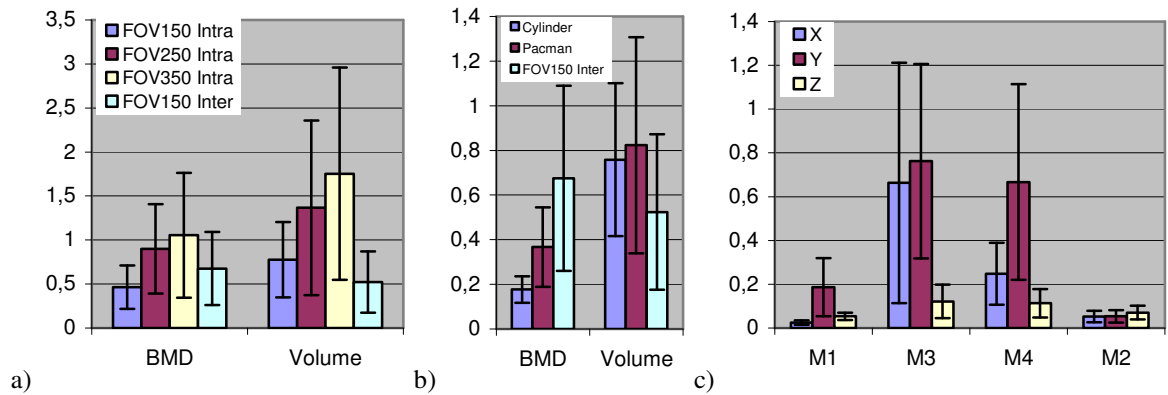

**Figure 2:** Precision errors as %CV for: a) Intra- and inter-operator study, b) Inter-operator study for the geometric VOIs, c) Inter-operator evaluation of the landmarks that define the VCS

**Precision.** The intra- and inter-operator precision analysis results are shown in Figure 2. Inter-operator precision was carried out for FOV=150 mm, only. Precision errors are given for trabecular BMD and total volume. The precision errors for BMD and volume for the geometrically defined VOIs (Figure 2b) are below 0.5 % and 1 %, respectively which is an excellent result. In Figure 2c precision errors for the positions of the 4 landmarks defining the VCS are given.

Our procedure takes up to 15 minutes with a trained operator on a standard Pentium IV PC with 2.8 Ghz and 1 GB of main memory for the assessment of L1-L3.

## 4 Conclusions

Our multi step approach has resulted in excellent segmentation results. An initial analysis of the new 3D segmentation method demonstrated lower precision and accuracy errors than existing methods [2]. We have shown that a multi-step procedure is an effective tool for 3D segmentation. The combination of several steps corrects deficiencies of the individual methods. Improved performance should result in superior diagnostic performance which is very important for fracture prediction in osteoporosis.